\documentclass[lnbip]{svmultln}
\usepackage{graphicx}
\usepackage{tabularx}
\usepackage{multirow}
\usepackage{array}

\usepackage{booktabs,subcaption,amsfonts,dcolumn}
\usepackage[dvipsnames]{xcolor}
\graphicspath{{images/}}

\begin{document}
\title{ProtoNER: Few shot Incremental Learning for Named Entity Recognition using Prototypical Networks}

\titlerunning{Few Shot Incremental Learning for NER}
\author{Ritesh Kumar \and Saurabh Goyal \and Ashish Verma\footnote{Work done while author was working at IBM Research.} \and Vatche Isahagian }
\authorrunning{Ritesh Kumar et al.} 
\tocauthor{kumar.ritesh@ibm.com, saurabh.goyal1@ibm.com, draverma@amazon.com, vatchei@ibm.com}
\institute{IBM Research, USA \\
\email{kumar.ritesh@ibm.com}, \email{saurabh.goyal1@ibm.com}, \email{draverma@amazon.com}, \email{vatchei@ibm.com}}

\maketitle    

\begin{abstract}      

Key value pair (KVP) extraction or Named Entity Recognition(NER) from visually rich documents has been an active area of research in document understanding and data extraction domain. Several transformer based models such as LayoutLMv2\cite{lmv2}, LayoutLMv3\cite{lmv3}, and LiLT\cite{LiLT} have emerged achieving state of the art results. However, addition of even a single new class to the existing model requires (a) re-annotation of entire training dataset to include this new class and (b) retraining the model again. Both of these issues really slow down the deployment of updated model.   \\
We present \textbf{ProtoNER}: Prototypical Network based end-to-end KVP extraction model that allows addition of new classes to an existing model while requiring minimal number of newly annotated training samples. The key contributions of our model are: (1) No dependency on dataset used for initial training of the model, which alleviates the need to retain original training dataset for longer duration as well as data re-annotation which is very time consuming task, (2) No intermediate synthetic data generation which tends to add noise and results in model's performance degradation, and (3) Hybrid loss function which allows model to retain knowledge about older classes as well as learn about newly added classes.\\ 
Experimental results show that ProtoNER finetuned with just 30 samples is able to achieve similar results for the newly added classes as that of regular model finetuned with 2600 samples. 

\end{abstract}

\section{Introduction}
Business processes provide a structured framework for enterprises to do work. They define tasks, their executors, and capture dependencies as well as provide logging and tracking capabilities  \cite{Weske2012Business}. They also align with company policies and compliance with governmental regulations.
Business process tasks are typically associated with unstructured data in the form of documents, which contain information deemed critical to the successful execution of the business process. For example, a loan application will be associated  with multiple documents containing name, salary, credit score etc. of an individual. In the age of digital transformation, where enterprises are focusing on augmenting business processes with Artificial Intelligence \cite{rizk2020conversational, Rizk2022Can, Huo2021Graph}, automating the extraction of knowledge from these rich documents such as loan applications, invoices, purchase orders, and utility bills, understanding business documents is critical because Incomplete, and Inaccurate information can lead to process execution delay and loss of revenue. Most recently, key-value pairs extraction has received significant attention because of its ability to influence the automation of several downstream tasks and affect the completion time of the business processes.

Traditional approaches such as template matching and region-segmentation based models \cite{regionBased1, regionBased2, heuristicBased1, heuristicBased2} have been commonly used in industry for KVP extraction as they provide flexibility to train and deploy the models at much faster pace. Unfortunately, these models only work for the documents that they have observed during model training time and even a slight change in the layout of the document results in poor performance \cite{templateMatching}.

Deep learning based models such as LayoutLMv3 \cite{lmv3} and FormNet \cite{FormNet} achieve state of the art results and work very well even for the unseen documents. Such properties of deep learning models have compelled their industry wide rapid adoption. Unfortunately, these models are not able to predict a new set of key classes for which the model is not explicitly trained. With the ever-evolving nature of form like documents, it becomes crucial for such models to support addition of new key classes on top of existing ones in a fairly simplified and straight forward manner.
With the goal of addressing this issue, in this paper, we adopt Prototypical Network\cite{ProtoNet} based model architecture to support the addition of new key classes to an already trained model. Prototypical networks have been widely studied for several computer vision related tasks but their inclusion in language related tasks remains limited. This is mainly due to the fact that several existing machine learning based algorithms can be used to extract features from the images pertaining to a class and be treated as prototype whereas there is not a clear or equivalent approach readily available that can be exploited to create prototypes in the language domain. We present a novel approach to create prototypes corresponding to different classes in the language domain that uses only a few samples to facilitate few shot class-incremental learning(FSIL), while avoiding the model's Catastrophic Forgetting \cite{CatFor} problem.

\section{Problem Formulation}

\begin{figure}[!h]
\vspace{-0.1in}
\centering
\frame{\includegraphics[width=0.90\linewidth]{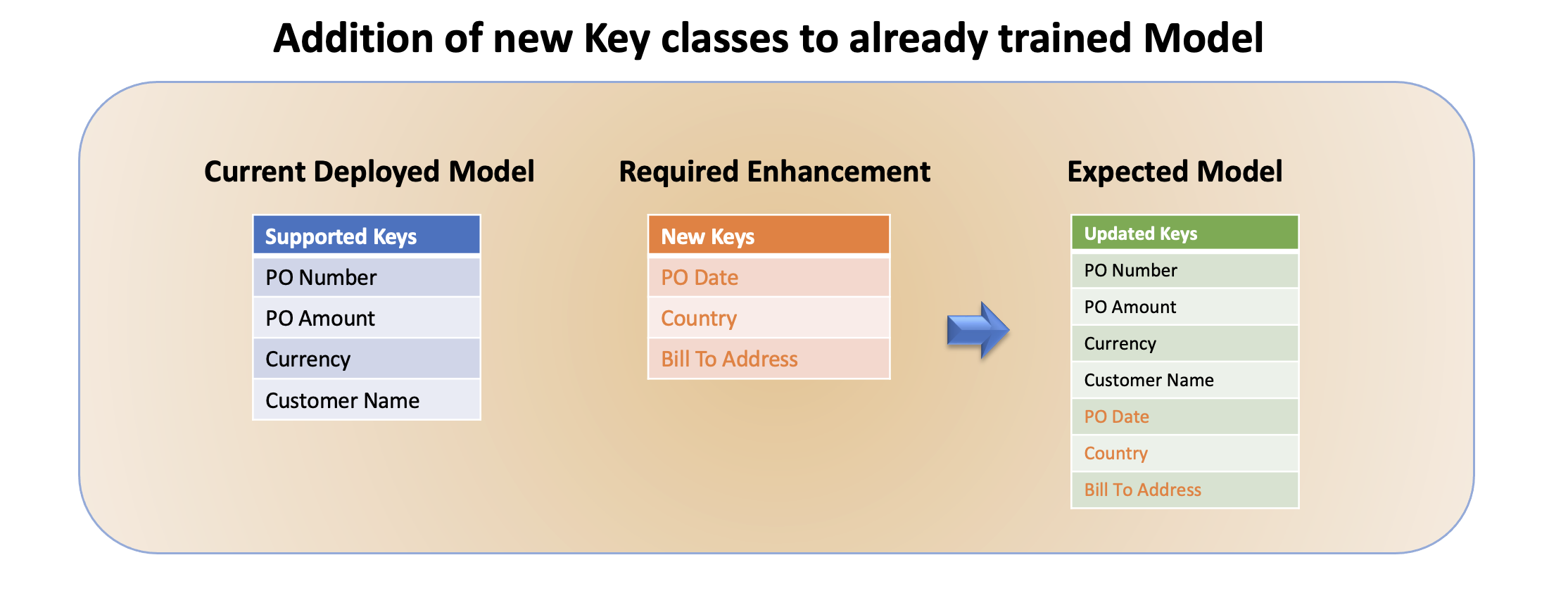}}
    \caption{\small Current deployed model supports extracting 4 classes. How to enhance the understanding of the model to support three additional classes.}
    \label{prob_stmnt}
\end{figure}

As described earlier, with rapid digitization of business workflows, enterprises are expected to update their models frequently with the capability of extracting more and more key value pairs from visually rich documents. The problem can be illustrated using  Figure \ref{prob_stmnt}. Consider model $M_0$ trained to extract 4 key classes (PO Number, PO Amount, Currency, and Customer Name) is currently being used by an enterprise to automate several downstream tasks (e.g. 3-way matching \cite{bpi2019}). However, the enterprise needs to further extract 3 more classes (PO Date, Country, and Bill To Address) along with the previous 4 keys to further extend the functionality. Therefore, the resulting new model $M_1$ needs to support 7 key classes in total instead of just 4. 
Formally, we define the problem as follows:

\begin{definition}
Given a set of $N$ classes, a Model $M_k$ is trained on $K$ classes where $K \subset N$. Train a new model $M_{k+j}$ trained on $K + J$ classes, where $J \subset N$, $K \cap J = \emptyset$, and $|J| + |K| \leq |N|$.
\end{definition}

\vspace{-0.1in}

In a conventional training setup, adding new classes requires re-annotation of the entire dataset to incorporate these new classes followed by retraining the model, which is a very slow and time consuming process because of its reliance on human annotators. This is further complicated by the fact that the original training dataset may not be available due to the data retention policies adopted by the organization. Our approach utilizes a handlful of newly annotated samples to facilitate few shot class-incremental learning(FSCIL)\cite{FSCIL}. 

Note that adding new classes to an already trained model may introduce a Catastrophic Forgetting\cite{CatFor} problem where model tends to forget the knowledge acquired about older classes while learning about new classes. To alleviate this problem, we incorporate a  hybrid loss function that combines cross entropy loss and cosine similarity loss. Cosine similarity loss is only applied to the older classes during addition of new class which forces the model to retain the knowledge about older classes whereas cross entropy loss allows the model to learn about new classes at the same time. It also helps the network not to overfit on the few-shot instances as well as not becoming biased to the base classes.

\section{ProtoNER} 

\subsection{Model Architecture} 
Transformer based NLP models such as LayoutLMv2\cite{lmv2} and LayoutLMv3\cite{lmv3} have shown to achieve state of the art results for KVP extraction on public datasets like CORD\cite{CORD} and FUNSD\cite{FUNSD} by leveraging text, layout, and image modalities. In this work, we leverage the LayoutLMv2\cite{lmv2} model architecture as the basis for our modified prototypical network based architecture. Note that our architecture is generic enough to support other multi-modal architectures that are capable of performing KVP extraction such as DocFormer \cite{DocFormer} or  TiLT \cite{TiLT}.

\begin{figure}[!h]
\centering
\vspace{-0.1in}    \frame{\includegraphics[width=0.90\linewidth]{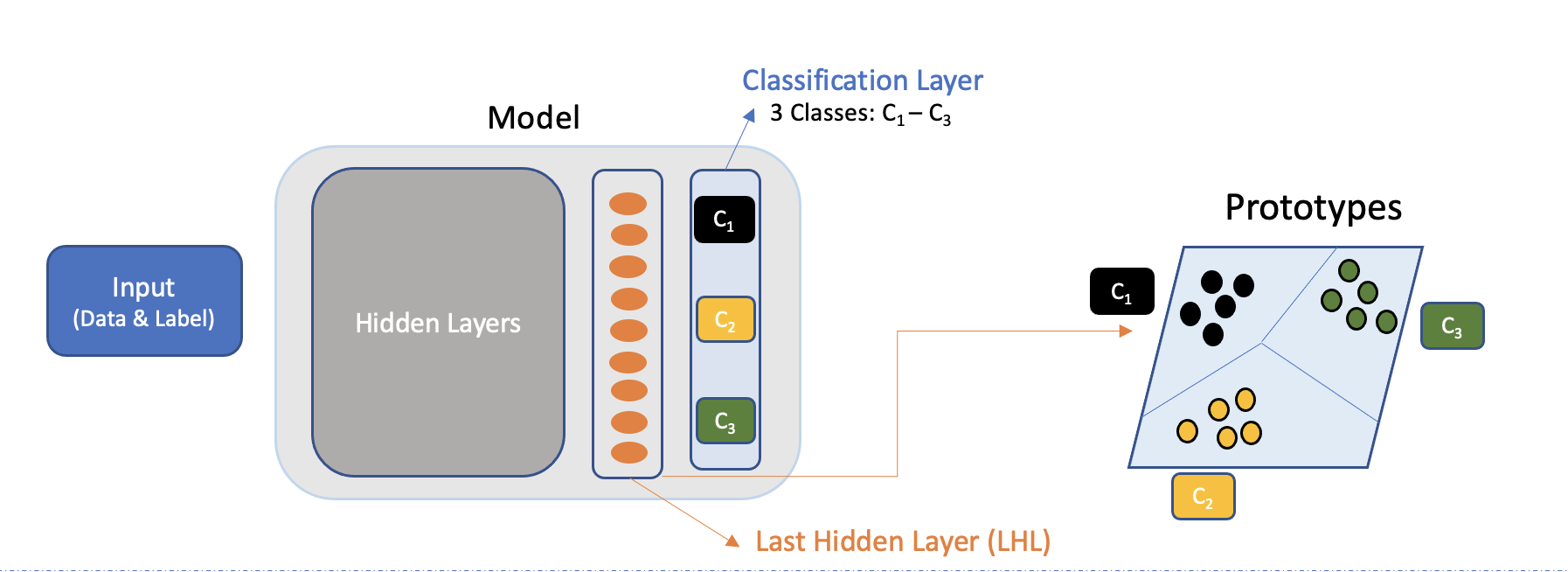}}
    \caption{\small Converting LayoutLMv2 to Prototypical Network Architecture} 
    \label{Model_architecture}
    \vspace{-0.2in}
\end{figure}

Figure \ref{Model_architecture} illustrates the overall architecture. For the sake of simplicity, all the transformer model blocks/layers are encapsulated as  ``Hidden Layers" block and only the last hidden layer(LHL) and classification layer are depicted.
Though LayoutLMv2 model supports sequence length of up to 512 tokens and classifies each token, we have shown the classification head only for one token here and named it as ``classification layer".

To convert the LayoutLMv2 model into prototypical network architecture, last hidden layer of the model is leveraged during training as well as inference. While training the model, if a token with ground truth label $C_1$ gets classified correctly by the classification layer, the last hidden layer representation for that token is saved as a prototype for key class $C_1$ under the prototype pool as shown in figure \ref{Model_architecture}. In a similar fashion, prototypes corresponding to all of the classes are saved during training. Multiple prototypes per class are saved to capture better diversity within the class prototypes. The number of prototypes per class to be saved is a hyper parameter. 
The prototypes are saved only during the last epoch of the training to allow the model learn and achieve good accuracy across all key classes before 
saving the prototypes. The prototypes are only saved when the model classifies the token correctly i.e the ground label for the token matches the predicted label. Each prototype is a vector of length equal to the length of the model's last hidden layer.

When the trained model is used for inferencing, the cosine similarity score is computed between the LHL representation of the token against all the saved prototypes from the pool and the label for the given token is derived by performing K-Nearest Neighbour search based on the computed cosine similarity scores. Doing the K-Nearest neighbour search from prototypes pool completely eliminates the need to have a classification layer.

\subsection{Training Procedure}

\begin{figure}[!htp]
\vspace{-0.2in}
\centering
    \frame{\includegraphics[width=0.85\linewidth]{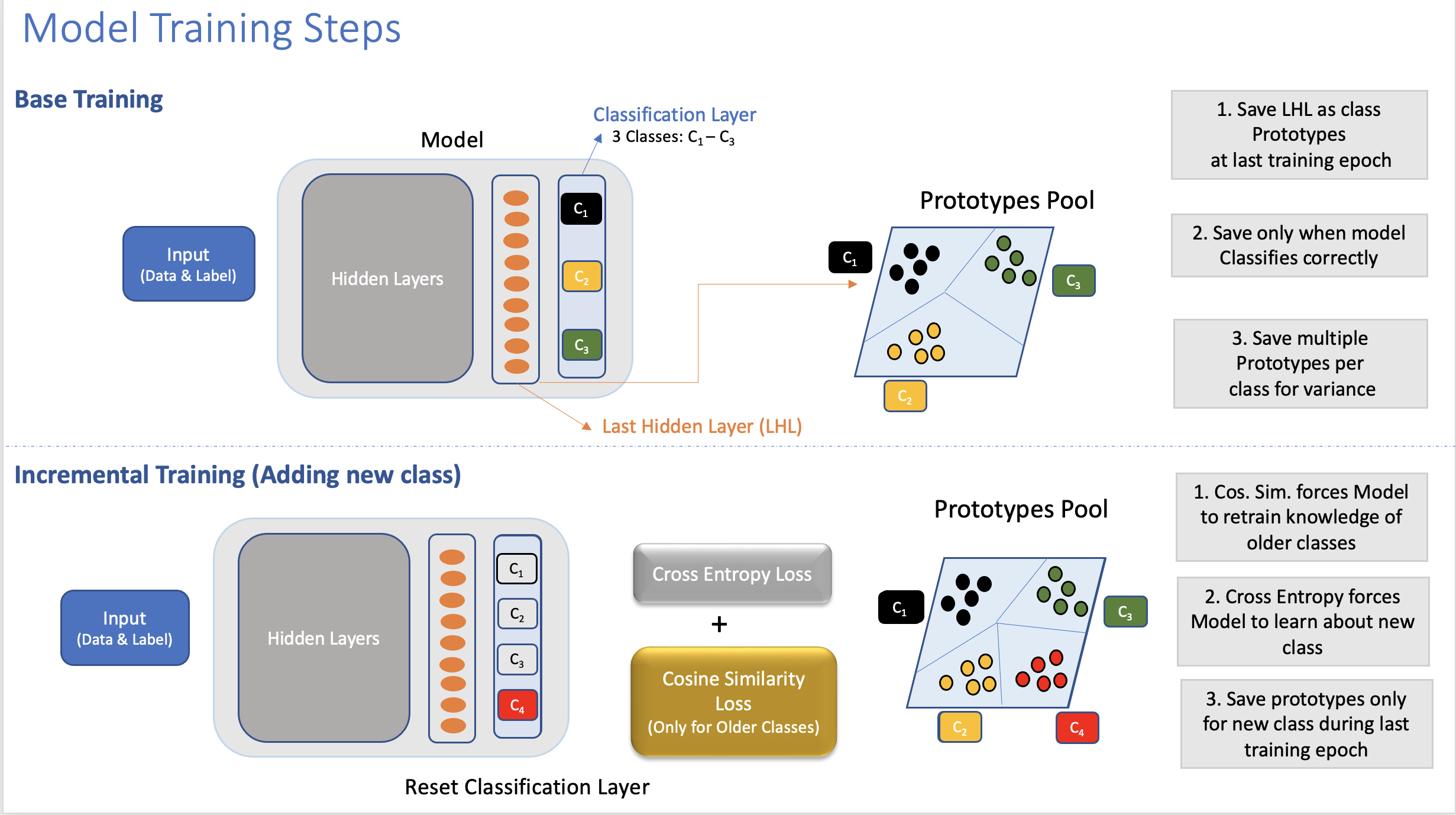}}
    \caption{\small Overall Model Training Process} 
    \label{model_training}
    \vspace{-0.2in}
\end{figure}

Figure \ref{model_training} highlights the overall training procedure for our model, which consists of two steps: Base Training, and Incremental Training. Base Training consists of training the model for the first N classes (N+1 if we include ``Other" class) and incremental training consists of adding new class to the model achieved after base training.\\
During the base training, the model is trained for N classes as per the procedure described in the previous section. The trained model weights achieved after the base training is termed as base model. Multiple prototypes are saved for all of the N classes to create the prototypes pool.\\
During the incremental training phase to add new class to the base model, the prototypes saved for the older N classes are carried forward as is and not updated at any point of time during the incremental training. Only the prototypes pertaining to the ``Other" class are discarded before initiating the incremental training. The classification layer of the base model is reset to reflect N+1 classes.  Using only the few newly annotated samples ( annotated for n+1 key classes) and trained base model weights, the model is finetuned further to acquire knowledge about this newly added class. Our hybrid loss function is used only during this incremental training phase. For the tokens pertaining to the older key classes, the cosine similarity loss is computed between the LHL representation for the token and prototypes for the respective key class from the prototypes pool. Since multiple prototypes per key class are saved, we compute the cosine similarity loss between LHL representation of the token and each prototype for that key class to derive the average loss. This loss is simply added linearly to the cross entropy loss computed for the same token. Other possibility could have been to associate learnable parameters with both of the losses and let the model learn them during training. We did not explore this possibility in this work.  \\
This way the model is forced to retain LHL representation for the older key classes as similar to the original state as possible. For the tokens pertaining to newly added key class, only the cross entropy loss is computed. Multiple prototypes are saved during the last epoch for the newly added class as well as the ``Other" class. The same procedure can be repeated again in the future to add additional key classes. Note that more than one new class can be added concurrently during the same incremental training phase\footnote{The addition of multiple classes sequentially (one at a time) vs. all at same time results in similar accuracy.}.

\section{Experimental Results}

\subsection{Dataset}
\begin{figure}[!htp]

\vspace{-0.2in}
\centering
    \includegraphics[width=0.8\linewidth]{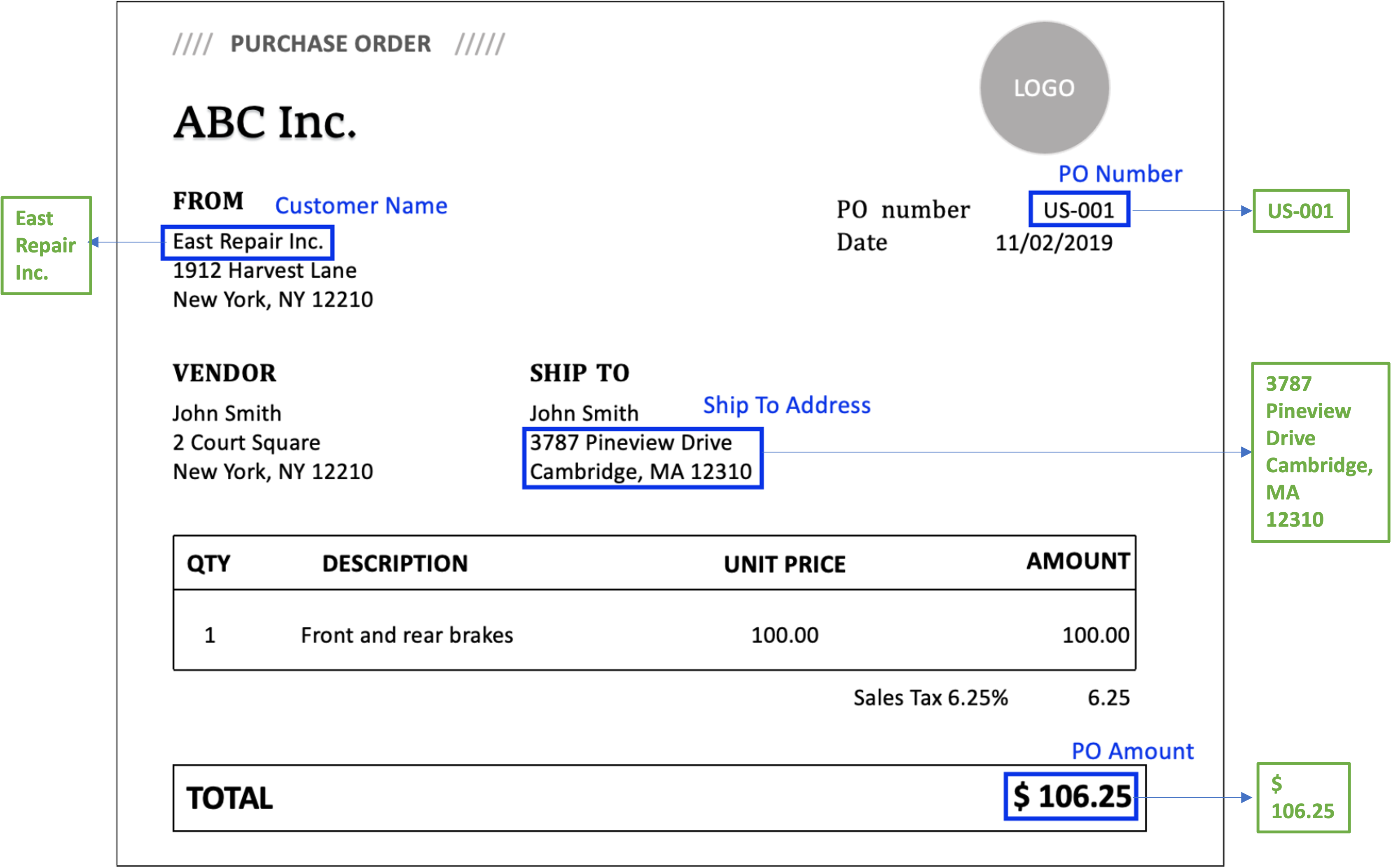}
    \caption{\footnotesize Purchase order sample showing annotated field level key labels (i.e. name, and location) of  Customer Name, PO Number, Ship To Address, and PO Amount along with their values. The list of text in the green boxes represents the word level OCR output obtained for each of the field level annotations.}
    \label{po_example}
    \vspace{-0.3in}
\end{figure}

The dataset used for the analysis contains 2742 purchase orders obtained from various sources and consists of about 73 unique layout templates. 2600 samples are used for training and 142 for evaluation.
Each document is annotated with a subset of the 10 pre-defined key classes. Table \ref{kvp_dataset} lists these key classes along with their respective frequencies i.e. how many times these key classes appear in the dataset. The annotations contain 2-D coordinates and the key class label for the values corresponding to the pre-defined keys (not the words corresponding to keys themselves). Since any document such as purchase order or invoice generally contains additional text that does not pertain to any of the key classes, we also include an ``Other" class along with the 10 pre-defined key classes to refer to those remaining words in the document. Note that all the annotations are at field level rather than at the word level as illustrated in Figure \ref{po_example}.

\begin{table}
    \footnotesize
    \centering
        \begin{tabular} {|c| l | c|} 
             \hline
             \bf \# & \bf Key Name & \bf Frequency  \\ [0.5ex] 
             \hline
             1 & PO NUMBER & 2377  \\
             2 & PO AMOUNT & 1384  \\
             3 & CUSTOMER NAME & 1168  \\
             4 & COUNTRY & 1033  \\
             5 & CURRENCY & 1311  \\
             6 & BILL-TO ADDRESS & 1334  \\ 
             7 & BILL-TO CUSTOMER NAME & 1030  \\
             8 & SHIP-TO ADDRESS & 1390  \\
             9 & SHIP-TO CUSTOMER NAME & 1050  \\
             10 & LOGO CUSTOMER NAME & 1631  \\[1ex] 
             \hline
        \end{tabular}
    \caption{\small \label{kvp_dataset}Frequency of keys present in Purchase Order dataset}
\end{table}

In order to process the data, each document is first passed through an Optical Character Recognition (OCR) engine to extract the words and their respective bounding box coordinates. 
Since annotations are done at field level and OCR extracts the text and corresponding bounding boxes at word level, we split the annotations at word level to align it with the OCR output. Both OCR output and pre-processed annotations are required for training the model for KVP extraction.\\

\subsection{Evaluation}
In order to evaluate our ProtoNER model, we follow the 2-step procedure  described in the earlier section to train our model. In the first step, we train the pre-trained LayoutLMv2 \cite{lmv2} model  on 2600 training samples annotated with only 4-classes as shown in Table \ref{tab:ProtoResults}. In order to do that, we modify the original training data annotations and replace all the key classes except the original 4 key classes with the ``Other" class. The model is trained for 100 epochs with 2e-5 learning rate and 8 batch size. At the end of this first step of training, we save 50 prototypes for each of the 4 key classes along with the model weights. The decision to save 50 prototypes per class was taken based on the empirical analysis.\\
In second step, we further fine-tune the model trained in the previous step with just 30 samples annotated with all the 10 key classes. The training parameters used for this step are: train epochs=100, learning rate=5e-6, and batch size =8. \\
This 2-step training regime mimics the practical industrial scenario where the initial model is usually trained on large dataset with small number of key classes and the model needs to be updated to cater for future requirements i.e. identify new key classes but with limited training data.

\begin{table}
\vspace{-0.1in}
    \begin{subtable}[t]{0.48\linewidth}
        \begin{tabular} {l   r} 
          \bf Base Model Training  \\
            \hline
            Train Samples &   2600 \\
            Keys & 4\\
            Prototypes per class & 50 \\
            Test samples & 142\\
        \end{tabular}
        \label{tab:ProtoResults_a}
    \end{subtable}
    \hspace{\fill}
    \begin{subtable}[t]{0.48\linewidth}
    \flushright
    
        \begin{tabular} {l   r } 
          \bf Incremental Model Training  \\
            \hline
            \color{Red} Train Samples &   \color{Red}30 \\
            \color{Red} Keys & \color{Red}6\\
            Prototypes per class & 50 \\
            Test samples & 142\\
        \end{tabular}
        \label{tab:ProtoResults_b}
    \end{subtable}
    \bigskip 
    
    \begin{subtable}[t]{0.48\linewidth }
        \begin{tabular}{| l | l | l | l |}
            \hline
            \textbf{Key Class} & \textbf{Prec } &  \textbf{Rec } &  \textbf{F1} \\
            \hline
            \color{Blue} PO Number           & 0.87 & 0.79 & 0.83 \\
            \color{Blue} Logo Cust Name      & 0.87 & 0.79 & 0.83 \\
            \color{Blue} Ship To Addr        & 0.83 & 0.85 & 0.84 \\
            \color{Blue} Ship To Cust Name   & 0.76 & 0.87 & 0.81 \\
            \hline
        \end{tabular} 
        \caption{\footnotesize \textbf{Base model accuracy}}
        \label{tab:ProtoResults_c}
    \end{subtable}
    \hspace{\fill}
    \begin{subtable}[t]{0.48\linewidth}
    \flushright
        \begin{tabular}{| l | l | l | l |}
            \hline
            \textbf{Key Class} & \textbf{Prec } &  \textbf{Rec } &  \textbf{F1} \\
            \hline
            \color{Blue} PO Number           & 0.87 & 0.89 & 0.88 \\
            \color{Blue} Logo Cust Name      & 0.90 & 0.76 & 0.81 \\
            \color{Blue} Ship To Addr        & 0.87 & 0.78 & 0.82 \\
            \color{Blue} Ship To Cust Name   & 0.76 & 0.88 & 0.81 \\
            Bill To Addr        & 0.63 & 0.85 & 0.72 \\
            Bill To Cust Name   & 0.74 & 0.79 & 0.76 \\
            Country             & 0.68 & 0.86 & 0.76 \\
            Currency            & 0.68 & 0.90 & 0.78 \\
            Customer Name       & 0.67 & 0.73 & 0.69 \\
            PO Amount           & 0.76 & 0.77 & 0.76 \\
            \hline
        \end{tabular}
        \caption{\footnotesize \textbf{Incremental model accuracy}}
        \label{tab:ProtoResults_d}
    \end{subtable}

\caption{\footnotesize ProtoNER training samples, keys, and model results are shown for base training (under subtable (\subref{tab:ProtoResults_c})) and incremental training (under subtable (\subref{tab:ProtoResults_d})). Common key classes across both models are highlighted in blue color.}
\label{tab:ProtoResults}
\vspace{-0.1in}
\end{table}

Table \ref{tab:ProtoResults} illustrates the Precision, Recall, and F1 scores achieved for key classes by \textbf{ProtoNER}. Table \ref{tab:ProtoResults_c} shows the scores for the base model trained with 4 key classes and Table \ref{tab:ProtoResults_d} shows the scores for the base model finetuned further to support 10 key classes. From the Table \ref{tab:ProtoResults}, it can be observed that the model is able to perform well for the newly added keys even after training with only 30 newly annotated samples. The hybrid loss function is able to force the model to retain the knowledge about older classes as well as gain knowledge about new classes. Also, the scores improve for the older keys after addition of new keys. The reason behind this improvement is that the false positives and false negatives for the 4 key classes get spread over 10 key classes now instead of 4.\\
The rationale behind how model is able to learn about new key classes from only few samples can be attributed to the sub-clustering being performed by the model inherently during the base training itself. Even though the words pertaining to left out 6 key classes are labeled as ``Other" during the base training, the model inherently forms sub-clusters under the parent ``Other" class umbrella for these 6 key classes. Exposing the model with few samples containing new key classes during incremental stage allows the mapping of such already formed sub-clusters to these new key classes. Meihan et al. \cite{otherClass} have reported similar observations under their few shot work.\\

\begin{table}
\vspace{-0.1in}
    \begin{tabularx}{1.03\textwidth}{|m{11em} m{2em} | m{8.5em} m{2em} | m{11em} m{2em} | }
        \multicolumn{6}{c}{ \textbf{Training Attributes}}\\
        \hline
        \multicolumn{2}{|c}{ \color{Blue}ProtoNER} &
        \multicolumn{2}{|c}{ \color{Blue}LayoutLMv2-10C} &
        \multicolumn{2}{|c|}{ \color{Blue}LayoutLMv2-4C-10C} \\
        \hline
        Base train samples & 2600 & Train samples & 2600 & Train samples & 2600 \\
        Base key classes & 4 & Base key classes & 10 & Base key classes & 4 \\
        Test samples & 142 & Test samples & 142 & Test samples & 142 \\
        \color{Red}Incremental samples & \color{Red}30 & & & \color{Red}Incremental  samples & \color{Red}30 \\
        \color{Red}Incremental key classes & \color{Red}6 & & & \color{Red}Incremental key classes & \color{Red}6 \\
        \hline
    
    \end{tabularx}
    \caption{\footnotesize Training configuration for ProtoNER, LayoutLMv2-10C, and LayoutLMv2-4C-10C  models. Both ProtoNER and LayoutLMv2-4C-10C are first trained for 4 key classes using 2600 samples followed by incremental training for additional 6 classes using 30 samples. }
    \label{tab:train_attrib}
\end{table}

\begin{table}
\vspace{-0.2in}
\setlength{\tabcolsep}{4pt}
    \begin{tabular}{ |l|| c| c| c|| c| c| c|| c| c| c|| }
        \multicolumn{10}{c}{ \textbf{Results  Comparison}}\\
        \hline
         & 
        \multicolumn{3}{|c||}{ \color{Blue} \footnotesize ProtoNER} &
        \multicolumn{3}{|c||}{ \color{Blue} \footnotesize LayoutLMv2-10C} &
        \multicolumn{3}{|c||}{ \color{Blue} \footnotesize LayoutLMv2-4C-10C} \\
        \hline
        \textbf{Key Classes}& \textbf{Prec} & \textbf{Rec} & \textbf{F1} & \textbf{Prec} & \textbf{Rec} & \textbf{F1} & \textbf{Prec} & \textbf{Rec} & \textbf{F1}\\
        \hline
        PO Number           & 0.87 & 0.89 & 0.88 & 0.88 & 0.82 & 0.84 & 0.77 & 0.55 & 0.64\\
        Logo Cust Name      & 0.90 & 0.76 & 0.81 & 0.88 & 0.82 & 0.84 & 0.77 & 0.46 & 0.57\\
        Ship To Addr        & 0.87 & 0.78 & 0.82 & 0.84 & 0.81 & 0.82 & 0.45 & 0.42 & 0.43\\
        Ship To Cust Name   & 0.76 & 0.88 & 0.81 & 0.77 & 0.89 & 0.82 & 0.42 & 0.39 & 0.40\\
        Bill To Addr        & 0.63 & 0.85 & \color{Red}0.72 & 0.81 & 0.83 & \color{Red}0.81 & 0.39 & 0.50 & \color{Red}0.44\\
        Bill To Cust Name   & 0.74 & 0.79 & \color{Red}0.76 & 0.80 & 0.84 & \color{Red}0.81 & 0.43 & 0.49 & \color{Red}0.46\\
        Country             & 0.68 & 0.86 & \color{Red}0.76 & 0.82 & 0.85 & \color{Red}0.83 & 0.62 & 0.64 & \color{Red}0.62\\
        Currency            & 0.68 & 0.90 & \color{Red}0.78 & 0.80 & 0.93 & \color{Red}0.86 & 0.72 & 0.78 & \color{Red}0.75\\
        Customer Name       & 0.67 & 0.73 & \color{Red}0.69 & 0.73 & 0.79 & \color{Red}0.75 & 0.50 & 0.69 & \color{Red}0.58\\
        PO Amount           & 0.76 & 0.77 & \color{Red}0.76 & 0.86 & 0.88 & \color{Red}0.86 & 0.64 & 0.62 & \color{Red}0.62\\
        \hline
    
    \end{tabular}
    \caption{\footnotesize Results comparison between ProtoNER, LayoutLMv2-10C, and LayoutLMv2-4C-10C models obtained for 142 test samples. LayoutLMv2-10C model is trained for all 10 key classes using 2600 samples. LayoutLMv2-4C-10C and ProtoNER models are first trained for 4 key classes using 2600 samples followed by incremental addition of 6 key classes using only 30 samples.  F1 scores for incremental key classes for both ProtoNER and LayoutLMv2-4C-10C models are highlighted in red color.}
    \label{tab:results}
    \vspace{-0.2in}
\end{table}

\subsection{Comparison against LayoutLMv2 model}
We also trained 2 baseline LayoutLMv2 models to compare against our model. We used the original implementation source code provided by the authors of LayoutLMv2 here: \url{https://github.com/microsoft/unilm/tree/master/layoutlmv2}. The first baseline model \textbf{LayoutLMv2-10C} was trained by fine-tuning the pre-trained LayoutLMv2 model on 2600 training samples annotated with all the 10 key classes. The model is trained for 100 epochs with 2e-5 learning rate and 8 batch size.\\
The second baseline model \textbf{LayoutLMv2-4C-10C}  was trained in 2 steps. It was first trained for 4 key classes using 2600 samples followed by finetuning further for all 10 key classes using only 30 samples. It was trained for 100 epochs with 2e-5 learning rate and 8 batch size followed by finetuning for 10 key classes for 100 epochs, 5e-6 learning rate and 8 batch size. All the models were trained on single V100 GPU.

Table \ref{tab:train_attrib} lists the overall training configuration for all of the 3 models. Table \ref{tab:results} compares the precision, recall and F1-score for all the 10 key classes for our model(ProtoNER) with LayoutLMv2-10C baseline model. It can be observed that for the original 4 key classes (PO Number, ShipToAddr, ShipToCustName, and LogoCustName) the F1-score of our model is comparable to the LayoutLMv2-10C model with 3\% drop only for 1 specific key class(LogoCustName). For the remaining 6 key classes, our model is able to learn only with the help of 30 new samples. The results suggest that the hybrid loss function is able to force the model to retain the knowledge about the original 4 key classes during the incremental training phase as well as achieves about 90\% of the LayoutLMv2-10C model’s accuracy for the newely added key classes with just 30 samples.

Table \ref{tab:results} also shows the comparison of precision, recall and F1-score between our model(ProtoNER) and LayoutLMv2-4C-10C model. It can be observed that our model performs significantly better than the LayoutLMv2-4C-10C model on original 4-classes with gains in F1-score as high as 40\% for some of the key classes and 32\% on average. This is due to the fact that our model is able to retain the knowledge about the original 4-classes in the form of saved prototypes and hybrid loss function while LayoutLMv2-4C-10C model suffers from catastrophic forgetting problem. For the newly added 6 classes, layoutLMv2-4C-10C model undergoes severe over-fitting due to the small dataset size while our model is able to generalize better due to hybrid loss function.

\section{Related Work}
The adoption of language models (LM) really demands the flexibility of continual and incremental learning. In context of incremental learning for KVP/NER, Chen and Moschitti \cite{CL2019} present an approach for transferring knowledge from one model trained on specific dataset to a new model trained on another dataset containing new keys/classes. Their overall model architecture tries to learn the differences between the source and target label distribution with the help of neural adapter. Greenberg et al. \cite{MargLike} use marginal likelihood training to strengthen the knowledge acquired by their model from different available datasets while filling in missing labels for each dataset to align them. Both of these models require availability of more than one annotated datasets and also does not incorporate the few shot training aspect.  Huang et al. \cite{FSSurvey} present a comprehensive study on the few shot training for NER task and mention about noisy supervised approach, knowledge distillation based teacher student model and prototypical networks based model. \\
On one hand, incremental learning aspect is being explored to allow addition of new classes to already trained model, on the other hand, different Few Shots based techniques are being developed to train model with minimal number of annotated examples using transfer learning. However, very limited work has been carried which exploits both Few Shot and Incremental Learning at the same time specifically for NER task. \\
Monaikul et al. \cite{AmazonNER} present model for incremental learning for NER task which follows teacher student architecture. Their approach passes the data through trained base model and considers the predicted labels as the ground truth label during the incremental training phase. It adds impurities to the training data since all the inaccurate predictions by the base model get passed to the incremental training. The complex inference head also requires sophisticated rules to eventually derive the final prediction which may induce/lead to inaccuracies. Zhou et al.\cite{MultiPhase} present meta learning based approach which relies on synthetically generated data. The model presented by Cheraghian et al.\cite{FSIL} few shot class incremental learning corresponds to vision domain where each sample contains data point pertaining to only one class and therefore, this model cannot be leveraged as is for the KVP task.

\vspace{-0.2in}
\section{Conclusion}
Our approach demonstrates how a prototypical network architecture inspired model setup with hybrid loss function can be used to incorporate real-life constraints and still achieve similar results as that of regular model. It provides a solution to add new keys on top of already trained model if and when required in the future with very limited data. It also eliminates the need to retain the original training dataset that could be a challenge in real-life scenario due to data retention policy adopted by different organizations. The overall setup provides flexibility to deploy such models in automated environment where end user can decides to add new keys with significantly less efforts.
%
%

\end{document}